\documentclass[10pt,letterpaper]{article}

\usepackage{cogsci}

\cogscifinalcopy %%% Uncomment this line for the final submission

\usepackage[T1]{fontenc}
\usepackage[american]{babel}
\usepackage{csquotes}
\usepackage{newtxtext,newtxmath}  %%% TeX Gyre Termes
\usepackage{graphicx}
\usepackage{verbatim}
\usepackage{linguex}
\usepackage[
  backend=biber,
  style=apa,
  natbib=true,
  eprint=true,
  annotation=false,
]{biblatex}
\usepackage{float} %%% Roger Levy added this and changed figure/table placement to [H] for conformity to Word template, though floating tables and figures to top is still generally recommended!

\usepackage[hidelinks]{hyperref}

\usepackage[dvipsnames]{xcolor}

\newcommand{\km}[1]{\textcolor{DarkOrchid}{[KM: #1]}}
\newcommand{\pt}[1]{\textcolor{Cerulean}{[PT: #1]}}

\usepackage{mdframed}
\newmdenv[
  topline=false,
  bottomline=false,
  rightline=false,
  linecolor=gray,
  linewidth=1pt,
  skipabove=\topsep,
  skipbelow=\topsep
]{draftytext}

\newcommand{\HypEOneLCQ}{\textsc{TL;JustAsk}}
\newcommand{\HypEOneSCQ}{\textsc{NoNeedToAsk}}
\newcommand{\HypEOneLExh}{\textsc{TooManyToList}}
\newcommand{\HypEOneSExh}{\textsc{JustListThemAll}}
\newcommand{\HypETwoBinarized}{\textsc{IfUnsure,Ask}}
\newcommand{\HypETwoDoIt}{\textsc{JustDoIt}}
\newcommand{\HypETwoHCQ}{\textsc{WorthAsking}}
\newcommand{\HypETwoMessUp}{\textsc{Don'tMessUp!}}

\title{Act or Clarify? Modeling
Sensitivity to Uncertainty and Cost in Communication}

\author[1]{\mbox{Polina Tsvilodub}}
\author[1]{\mbox{Karl Mulligan}}
\author[1]{\mbox{Todd Snider}}
\author[2]{\mbox{Robert D. Hawkins}}
\author[1]{\mbox{Michael Franke}}
\affil[1]{Department of Linguistics, University of T\"ubingen (\texttt{first.last@uni-tuebingen.de)}}
\affil[2]{Department of Linguistics, Stanford University (\texttt{rdhawkins@stanford.edu})}

\begin{document}

\maketitle

\begin{abstract}
When deciding how to act under uncertainty, agents may choose to \textit{act to reduce} uncertainty or they may \textit{act despite} that uncertainty.
In communicative settings, an important way of reducing uncertainty is by asking clarification questions (CQs).
We predict that the decision to ask a CQ depends on both contextual uncertainty and the cost of alternative actions, and that these factors interact: uncertainty should matter most when acting incorrectly is costly.
We formalize this interaction in a computational model based on \textit{expected regret}: how much an agent stands to lose by acting now rather than with full information.
We test these predictions in two experiments, one examining purely linguistic responses to questions and another extending to choices between clarification and non-linguistic action.
Taken together, our results suggest a rational tradeoff: humans tend to seek clarification proportional to the risk of substantial loss when acting under uncertainty.

\end{abstract}

\section{Introduction}

%\mf{Intro outline by Michael:} 
%uncertainty about the context and when to act on it: acting under the uncertainty or reacting TO the uncertainty; standard Bayesian models assume a fixed world model, but sometimes we need to revise it --- we pick clari-qs because this is where we can actually observe humans act upon it. We provide a model: we motivate regret as being based on an independent measure of options being good enough (Expt from CPL + model), and then Expt 2 looks at more implication of the model. 

%\begin{itemize}
    %\item 
We regularly face uncertainty when deciding how to act---whether about the state of the world, the consequences of potential actions, or another person's goals and intentions.
When navigating such uncertainty, we face a fundamental choice: do we 
\textit{act despite that uncertainty}, accepting the risk of error, or do we first \textit{act to reduce the uncertainty}, actively seeking information before committing to action?
Both strategies have costs: acting under uncertainty risks mistakes, while seeking information takes time and effort and delays other action.
Which factors govern this tradeoff? 
%   \item clarification questions are an observable window into how people "reacting to" uncertainty
    %\item 

%\mf{the second paragraph doesn't really flow; drop first two sentences for more coherence w/o too much loss of content?}\ts{seconded}
%When information accumulates passively, an agent must choose the right time to commit to a decision and stop waiting for more information to arrive \citep[e.g., as in drift-diffusion models;][]{ratcliff2016diffusion}.

%A rational agent could resort to probabilistic reasoning when acting under uncertainty.
%Choosing actions, a rational agent will factor in the uncertainty through probabilistic reasoning \pt{some ref}.
A rational agent should consider the expected value of what she could learn when choosing to seek (possibly costly) information before acting, as prescribed, for instance, by statistical decision theory when contemplating whether to perform an experiment \citep[][]{RaiffaSchlaifer_1961:DecisionTheory}.
%The rational agent's choice to seek information actively at a cost will be governed by the expected value of the new information she could learn \citep[][]{RaiffaSchlaifer_1961:DecisionTheory}.
In communicative settings, however, humans have access to a distinctively social form of information-seeking: asking \textit{clarification questions} (CQs), 
%In communicative settings, \textit{clarification questions} (CQs) represent a distinctively social form of information-seeking. 
%Rather than sampling from the environment, 
to retrieve decision-relevant information from others
\citep{purver_2002, saparina2025reasoning,dong2026value}. 
Still, the decision of whether or which CQ to ask may be subject to similar rationality considerations as the choice of experiments or of other kinds of questions \citep{van2003questioning, benz2006utility, hawkins_2015, coenen2019asking, hawkins2025relevant, rothe2018people, van2003questioning}.
While many aspects of CQs have been investigated in prior work, such as the syntax of clarificational moves \citep{healey2003experimenting,purver_2002,purver_2003}, or how conversational language models might implement them \citep{andukuri2024stargate, dagan20252,kundu-etal-2020-learning,ma2025ambigchat,majumder-etal-2021-ask,testoni_2024}, the question of \textit{when} and \textit{why} human interlocutors choose to ask a CQ has received almost no attention in empirical or computational work \citep[][]{grand2026shoot}.

A prominent use of CQs is as a repair strategy for when an agent cannot fully resolve an utterance's content, helping to establish shared understanding \citep[or \textit{grounding};][]{ginzburg_1996, ginzburg_2012, clark_1989}. 
But CQs can also serve other functions \citep{schlangen_2004}, including resolving uncertainty about the context, such as when the current \textit{decision problem} (DP) is not optimally addressable due to missing information \citep[cf.~][]{zhang2024clamber}.
In such cases, rational choice theory makes two predictions:
CQs should be more likely when (i) uncertainty about decision-relevant features is high, and (ii) available actions have low expected payoff. 
Crucially, these factors should interact: uncertainty matters most when mistakes are costly. 
We propose that this interaction arises because both factors are naturally unified through \textit{expected regret} \citep[equivalently: the \textit{expected value of perfect information};][]{RaiffaSchlaifer_1961:DecisionTheory}: agents ask CQs  when their best available action risks substantial loss relative to what full information would afford. 
We test these predictions in a question-answering context where responses are purely linguistic (Experiment 1), then ask whether the same tradeoff governs choices between clarification and non-linguistic action (Experiment 2), before formalizing the account in a layered computational model based on expected regret.\footnote{All code and data are available at \url{https://tinyurl.com/3z4awh4r}.}
%\pt{Maybe more literature: \citep[][]{andukuri2024stargate, grand2026shoot, kuhn2022clam, zhang2024clamber}.}

    \begin{figure*}[t]
    \centering
    \includegraphics[width=1\linewidth]{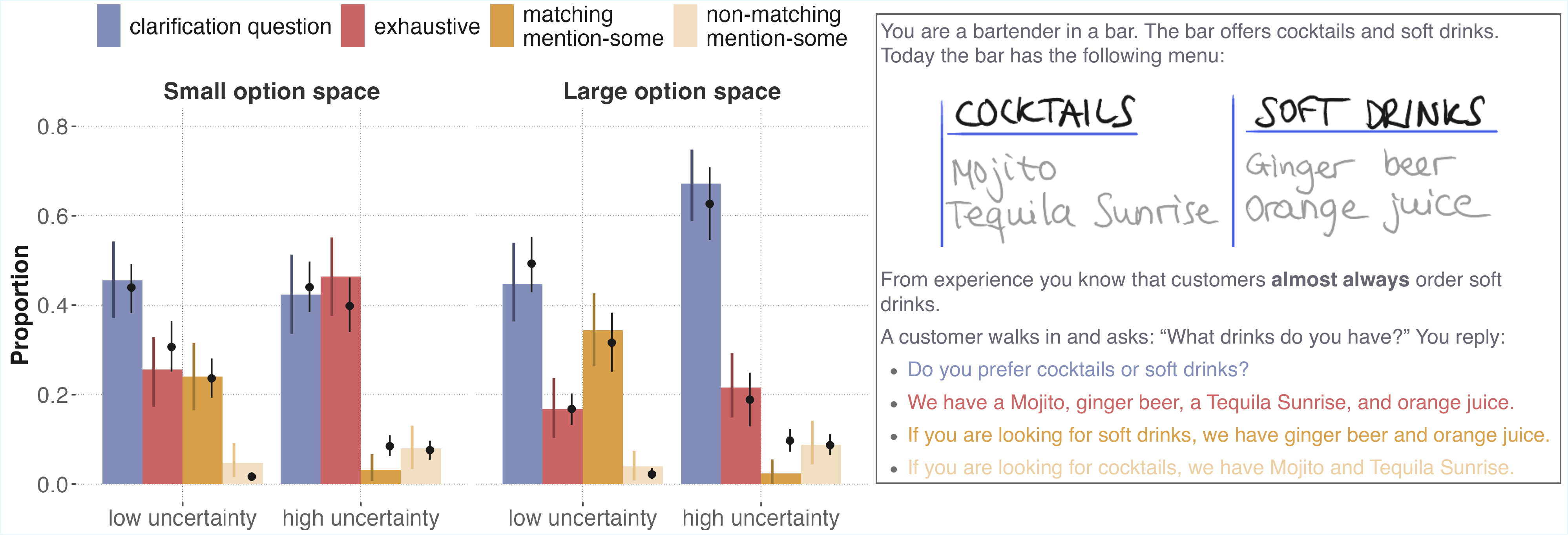}
    \caption{
    \textbf{Left}: Colored bars: Results of Experiment 1, showing proportions and 95\% bootstrapped CIs of selected responses (colors) in different uncertainty conditions (x-axis) under different costs 
    %of declarative answers 
    (facets).
    %In high uncertainty conditions, 
    Under high uncertainty,
    no preference is mentioned, so the mention-some labels were assigned to 
    %the 
    responses randomly.
    Black dots: posterior predictive means (dots) and 95\% credible intervals (black lines) of the computational model%
    %for each condition
    , under parameters fitted to human data via Bayesian data analysis. 
    \textbf{Right}: Example item from Experiment 1 in the small option space, low uncertainty condition. % \pt{I also have a version with means from alternative models, but it looks too busy IMO.}
%    \pt{how to add error bars on the human empirical data so that it is not confusing?}
  %  \mf{maybe use a darker shade of each CSP color for the empirical error bars; move the empirical ones slightly left, the model predictions slightly right, so that they both stay on the bar but are separate}
  %  \mf{looks like there is spurious white space underneath the figure; try adding this:   + labs(x = NULL, y = NULL) + theme(plot.margin = margin(t = 0, r = 0, b = 0, l = 0, unit = "pt")) }
    }
    \label{fig:e1}
\end{figure*}

\section{Experiments}
\label{sec:experiments}
In cooperative communicative contexts, we hypothesize that participants will trade off uncertainty about goal-relevant information against the cost of available actions when deciding whether to act despite uncertainty or to first ask a CQ.
We focus on two types of speech acts which signal the speaker's decision problem (or discourse goal): questions and directives.
We investigate the conditions which influence the propensity of interlocutors to ask a follow-up CQ, when there may be lingering uncertainty about the exact speaker goal.
In both cases, for questions and directives, cooperative actions are available without clarification, although they may carry high costs---e.g., giving an exhaustive answer to a question.

\subsection{Experiment 1: Reacting to Questions}
\label{sec:experiment-fc}

We begin with scenarios where someone asks a question (e.g., a customer walking into a bar, asking ``What drinks do you have?'') and an addressee aims to provide a helpful answer (e.g., the bartender offering options; see Figure~\ref{fig:e1}).
Previous work has shown that answerers reason about questioners' goals when selecting answers,
even while implicitly accepting uncertainty about those goals \citep[e.g.,][]{StevensBenz2016:Pragmatic-quest,hawkins2025relevant}. 
Building on this work, we investigate when answerers opt to \textit{reduce their uncertainty} by asking for clarification about the questioner's goal. 

More specifically, the answerer has uncertainty about the practical \textit{domain goal} that the question signals (e.g., getting a refreshing drink).
Whatever the domain goal, the answerer always has the option of providing an \textit{exhaustive} answer to the explicit question: a reliable but costly action.
Choosing a more efficient action (e.g., a \textit{mention-some} answer listing only non-alcoholic beverages) requires either information the answerer lacks or a willingness to act despite uncertainty about the questioner's practical goal.
Alternatively, posing a CQ to determine the questioner's goal may offer a route to an efficient answer, but asking carries its own costs. 
If CQ choices involve reasoning about uncertainty and the costs of alternative answers, we make the following exploratory predictions:
\begin{enumerate}
    \item \HypEOneLCQ:
        with a large option space, a higher rate of CQs under high uncertainty than under low uncertainty
    \item \HypEOneSCQ: 
        with a small option space, no difference in rate of CQs between high and low uncertainty
    \item \HypEOneSExh: 
        with a small option space, a higher rate of exhaustive answers under high uncertainty than low
    \item \HypEOneLExh:
        with a large option space, no difference in exhaustive answers between high and low uncertainty
\end{enumerate}

\paragraph{Methods \& Materials}
To investigate these hypotheses, we ran a $2 \times 2$ factorial  forced choice experiment, manipulating \textit{uncertainty} about the questioner's goal and the cost providing an answer. 
Uncertainty was operationalized through a cover story about the questioners' likely preferences (e.g., for drinks); cost was operationalized through the  number of contextually available options (\textit{option space} size).
Each trial listed specific options within two basic-level categories (e.g., cocktails and soft drinks), either two or four options per category (\textit{small} or \textit{large} option space). 
The cover story indicated that customers were equally likely to prefer either category (\textit{high} uncertainty) or almost certainly preferred one category (\textit{low} uncertainty; which category was counterbalanced). 
%\ex. You are a bartender in a bar. The bar offers cocktails and soft drinks. Today the bar has the following menu (presented in a picture in the experiment): Cocktails: Mojito, Tequila Sunrise. Soft drinks: Ginger beer, orange juice. 
%From experience you know that customers are equally likely to order cocktails and soft drinks.
%A customer walks in and asks: ``What drinks do you have?'' You reply: \label{ex:e1}
%    \a. Do you prefer cocktails or soft drinks? \OnRight{\textcolor{gray}{[clarification question]}}
%    \b. We have a Mojito, ginger beer, a Tequila Sunrise, and orange juice. \OnRight{\textcolor{gray}{[exhaustive answer]}}
%    \c. If you are looking for soft drinks, we have ginger beer and orange juice. \OnRight{\textcolor{gray}{[mention-some answer matching one goal]}}
%    \d. If you are looking for cocktails, we have Mojito and Tequila Sunrise. \OnRight{\textcolor{gray}{[mention-some answer matching other goal]}}

Participants ($N=125$) were recruited via Prolific, restricted to self-reported native English speakers in the US and UK, with approval rates over 95\% and at least five prior studies.
Each participant completed four trials (one per condition; within-subject) with items randomly sampled from a total of six.
Participants selected  among four response options (presented in randomized order). % as shown in the example. 
The experiment took approximately three minutes and participants were paid \pounds 0.45. 
An example trial is shown in Figure~\ref{fig:e1} (right).

\paragraph{Results}
Results are shown in Figure~\ref{fig:e1} (colored bars).
Visual inspection supports \HypEOneLCQ~and~\HypEOneSExh.
Additionally, when the option space is large but uncertainty is low, preference-matching mention-some answers are more common than under high uncertainty. 
To statistically test these qualitative patterns, we fit separate Bayesian logistic regression models for each answer type, with main effects of uncertainty, option space, and their interaction, with maximal random effects structure. All contrasts were sum-coded.%
    \footnote{The model in R syntax: \texttt{answer proportion $\sim$ uncertainty * space size + (1 | itemName) + (1 + uncertainty + space size | subject)}. No interaction in the by-subject RE was included due to convergence issues.} 
We report posterior means and 95\% credible intervals (CrIs excluding 0 indicate credible effects).

Confirming visual impressions, we found support for \HypEOneLCQ~($\beta = 1.58 [0.75, 2.48]$) and \HypEOneSExh~($\beta=2.00 [0.87,3.74]$), and, as predicted, no credible support for \HypEOneSCQ~($\beta=-0.21 [-0.99, 0.58]$) or \HypEOneLExh~($\beta = 0.99[-0.32,2.66]$).
We also found credible main effects of uncertainty and option space size on the CQ rates ($\beta=0.34 [0.04,0.67]$ and $\beta = 0.40 [0.13, 0.67]$, respectively) and on exhaustive answer rates ($\beta=0.75 [0.24, 1.51]$, $\beta = -0.68 [-1.24, -0.26]$).
Finally, mention-some responses were more frequent under low uncertainty with a large option space ($\beta = -7.56 [-18.80,-2.63]$) but not a small one ($\beta=-2.33 [-6.82, 1.54]$). 
These results provide 
empirical support for our hypothesis that communicators are sensitive to uncertainty (asking for clarification when uncertainty is high), and that this propensity
is modulated by costs of alternataive responses.

\subsection{Experiment 2: Reacting to Directives}
\label{sec:experiment-rating}
Experiment 1 provided evidence that people are sensitive to uncertainty, preferring CQs over other linguistic actions when costs are high.
In Experiment 2, we broadened our scope to include decision problems addressable by \emph{non-linguistic} actions. 
We focused on responses to directives. 
Both directives (e.g., ``Please open the door.'') and questions (e.g., ``Would you please open the door?'') can indicate the speaker's goals, but we focused on directives for reasons of pragmatic directness and syntactic simplicity.
The respondent always had the option to select a direct action that might satisfy the directive, without clarification.
%\pt{repetition / moved here from intro: We focus on two types of cooperative communicative contexts which signal a clear discourse goal \pt{the word discourse is a bit our of context; maybe: shared goal?}: questions and directives. For obvious reasons, the most cooperative reaction to a directive is completing the requested action. However, there are many cases where it is unclear how exactly to accomplish the requested goal \pt{potentially move to E2 intro}.
%    In such cases, one strategy might be to ask for more information, i.e, ask for clarification.}

We also investigated whether this sensitivity is binarized or gradient by modulating the amount of uncertainty in the context.
This distinction matters theoretically, as a binarized pattern would suggest a threshold-based decision process (asking a CQ whenever there is any uncertainty), while a gradient pattern would suggest sensitivity to the degree of uncertainty.
To test if different \emph{sources} of uncertainty were treated differently, we also varied whether the directive was underdetermined with respect to reference, manner, or deadline.
Incorporating insights from Experiment 1, we also modulated the relative cost of acting---here, through the cost of mistakes---to see if this impacts how likely people are to ask clarification questions. We formulate the following exploratory predictions:
\begin{enumerate}
    \setcounter{enumi}{4} 
    \item \HypETwoBinarized: binarized (i.e., non-graded) effect of uncertainty on CQs (by cost condition) 
    \item \HypETwoDoIt: with less uncertainty, more direct action
    \item \HypETwoHCQ: with higher costs for acting incorrectly, more CQs (across uncertainty conditions) 
    \item \HypETwoMessUp: with higher costs for acting incorrectly, less direct action 
\end{enumerate}

\begin{comment}
\pt{TODO: motivate directives}
\km{attempting to motivate directives:} 
We examine responses to directives, which explicitly encode a speaker's decision problem (though we note that decision problems may implicitly underlie other kinds of speech acts, like questions and assertions, making our insights about clarification questions here applicable beyond directives).
Based on our definition of clarification question, we expect participants to ask more clarification questions when a decision problem is not immediately actionable due to uncertainty.
With our design, we investigate to what extent the impulse to ask for clarification is binarized or gradient with the amount of uncertainty in the action space. 
We also see whether a higher cost of acting incorrectly encourages the asking of clarification questions.
\end{comment}
%
\paragraph{Methods \& Materials} 
To address these questions, we designed a $2 \times 3$ rating experiment, where we manipulated the amount of uncertainty in the action space (\textit{no}, \textit{low}, or \textit{high} uncertainty), as well as the relative cost of selecting an incorrect action (\textit{low} or \textit{high} cost). 
Cost was manipulated through the consequences of an error.
For example, opening the wrong cabinet is low cost while breaking down the wrong door with a sledgehammer is high cost.
Participants ($N=120$, 2 excluded for failing attention checks) read scenarios in which they were directed to do something, and then were asked to rate the likelihood that they would choose each of several (linguistic or non-linguistic) actions by using a slider bar ranging from ``very unlikely'' to ``very likely''.
    Both the participant instructions and scenario descriptions indicated that the directive-giver and the participant were aligned in their goals and in a social relationship such that issuing a directive would be natural, resulting in a shared decision problem for the director and respondent.
    In the \textit{low} and \textit{high} uncertainty conditions, the requested action (i.e., the way to resolve the shared decision problem) was underdetermined in terms of one of
    \textit{reference} (uncertainty about what an expression refers to),
    \textit{manner} (uncertainty about how an action is meant to be performed),
    or
    \textit{deadline} (uncertainty about the timeframe in which the action is meant to be performed).
Participants were presented 
a context
and
a set of options, %: one, two, or three actions (corresponding to no, low, or high uncertainty); a clarification question; a non-clarification question (``Why should I do that?''); verbal acceptance (saying, ``OK'' and doing nothing else); or inaction (doing nothing) 
%(see Example~\ref{ex:e2}). 
%Participants were asked to rate how likely they would be to choose option by using a slider bar ranging from ``very unlikely'' to ``very likely''.
%do we need more prose introducing this? 
%for instance 
as in the following referential example ($\{$no\,$\vert$\,low\,$\vert$\,high$\}$ uncertainty, low cost):%
\ex.\label{ex:e2} You are in a small kitchen with $\{$a single\,$\vert$\,two\,$\vert$\,three$\}$ cabinets, all of which are currently closed. Your friend says, "Get me a pot from the cabinet!"\\
    %Options:
        \textcolor{gray}{Action1/2/3:} You open $\{$the\,$\vert$\,the left/right\,$\vert$\,the left/middle/right$\}$ cabinet.
        \textcolor{gray}{CQ:} You ask, ``Which cabinet are the pots in?''
        \textcolor{gray}{NCQ:} You ask, ``Why should I do that?''
        \textcolor{gray}{Accept:} You say, “OK” and do nothing else.
        \textcolor{gray}{Inaction:} You do nothing.
%You are a construction worker working on a renovation project. You are in a room with three doors, all of which are currently closed and locked. There is also a sledgehammer in the room. The foreman says, ``Open the door!''	
%    \a. You break down the left door with the sledgehammer. \OnRight{\textcolor{gray}{[action 1]}}
%    \b.  You break down the right door with the sledgehammer. \OnRight{\textcolor{gray}{[action 2]}}
%    \c. You ask, “Which door?” \OnRight{\textcolor{gray}{[CQ]}}
%    \d. You ask, “Why should I do that?” \OnRight{\textcolor{gray}{[non-CQ Q]}}
%    \e. You say, “OK” and do nothing else. \OnRight{\textcolor{gray}{[acceptance]}}
%    \f. You do nothing. \OnRight{\textcolor{gray}{[non-action]}}
%\noindent \textbf{Reactions}: You break down the left door with the sledgehammer. (\textbf{$a_1$}) / You break down the right door with the sledgehammer. (\textbf{$a_2$}) / You break down the middle door with the sledgehammer. (\textbf{$a_3$}) / You ask, “Which door?” (\textbf{clari-q})/ You ask, “Why should I do that?” (\textbf{non-clari-q})/ You say, “OK” and do nothing else. (\textbf{verbal-accept}) / You do nothing (\textbf{non-action}).\ts{some questions commented here} 
%tns: would this be better as a bulleted list?
%tns: there's some non-uniformity wrt the slashes and spacing
%tns: there's some non-uniformity wrt italics and boldface here (condition names are italicized in the prose elsewhere; is this meant to parallel that, or to be different from that?)

To streamline participants' interpretation of the sliders in relation to one another, we presented a pie chart which dynamically updated as sliders were adjusted, representing the likelihood of each option; sliders were initialized at equal proportions, and each slider had to be clicked at least once before proceeding to the next trial.
The provided ratings (0--100) were analyzed raw, to make responses across uncertainty conditions (with 1--3 direct action options) comparable.
%to consider each of the options in relation to one another, we presented a pie chart which dynamically updated as sliders were adjusted, representing the likelihood of each option; sliders were initialized at equal proportions, and each slider had to be clicked at least once before proceeding to the next trial.% \pt{very cool and would be lovely to share bc Karl put so much effort and heart and soul into them; but for space reasons, we might cut here...}.\km{lol}
%\pt{to be integrated: the pie-chart does not, strictly speaking, force tradeoffs across sliders: people could pick values from 1-100 for all sliders (the only normalization that happens is purely visual, added with the goal of streamlining interpretation of the ratings across participants). We do it this way and don’t normalize values because the results would not be comparable across uncertainty conditions (1, 2, 3 direct action options).}

\begin{figure*}
    \centering
    \includegraphics[width=\linewidth]{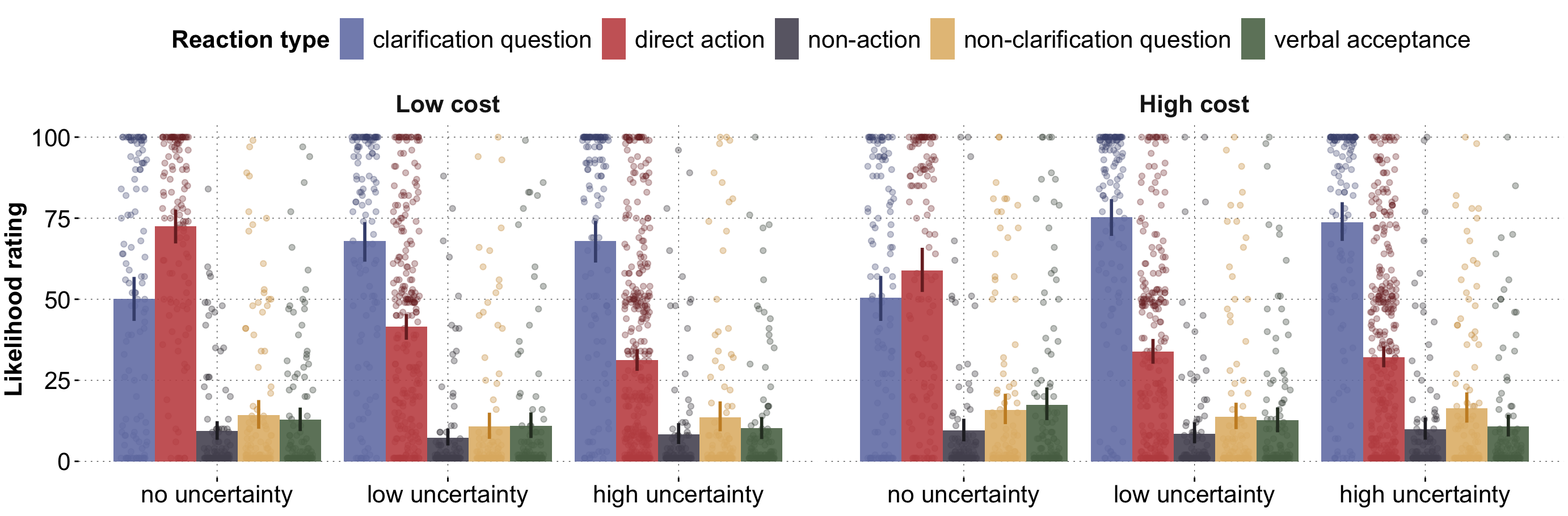}
    \caption{Results of Experiment 2: Unnormalized mean likelihood rating of different reaction options, grouped by uncertainty and costliness. Different direct actions were condensed into a single label for analysis.
        Error bars indicate 95\% bootstrapped CIs, and points represent individual trials.}
    \label{fig:e2}
\end{figure*}

\subsection{Results}

% \km{re-write in past tense}
We found effects of both uncertainty and cost on 
%various 
reactions to directives, including our main tradeoff of interest between asking CQs and taking direct action.
%The overall results are 
Results are 
%plotted 
in Figure \ref{fig:e2}. 

For statistical analysis, we fit a Bayesian linear regression model, regressing the ratings against the main effects of uncertainty, reaction type, cost, their interactions and random effects. 
All contrasts were sum-coded.\footnote{
The model in R syntax: \texttt{rating $\sim $ uncertainty * cost * reaction + (1 + uncertainty + cost $\vert \vert$ subject) + (1 + uncertainty + cost + reaction $\vert \vert$ item)}; maximal converging REs were used. All direct action options were pooled%
%into one category
.}
Starting with the effect of uncertainty, we found that across cost conditions, there was a credible effect of low uncertainty compared to no uncertainty on the rating of CQs ($\beta=19.90[14.79, 24.91]$), but no credible difference in the rating in the high compared to low uncertainty condition ($\beta=0.03[-4.99, 5.23]$), thus suggesting support for \HypETwoBinarized. 
% The decision to ask a CQ is binarized; it is \textit{not} gradiently sensitive to uncertainty.
% On the other hand, the likelihood of choosing direct actions decreased gradiently with increasing uncertainty (high~vs.~low: $\beta=-7.93, [-11.38, -4.43]$, low~vs.~no:$\beta=-25.85, [-30.42,-21.32]$), confirming intuitions about action under uncertainty. 
On the other hand, there was support for a gradient interpretation of \HypETwoDoIt~(high~vs.~low: $\beta=-7.93, [-11.38, -4.43]$, low~vs.~no:$\beta=-25.85, [-30.42,-21.32]$), confirming intuitions about hesitating to act under increasing uncertainty. 

% IfUnsure,Ask
We also found evidence for \HypETwoBinarized~both 
%when grouped by 
within
the high cost condition (right facet of Figure \ref{fig:e2}; low~vs.~no: $\beta=21.21 [15.72, 26.59]$, but not high~vs.~low: $\beta=0.68 [-4.75, 6.25]$) 
and the low cost condition (left facet of Figure \ref{fig:e2}; low~vs.~no: $\beta=18.60 [13.25, 24.00]$, but not high~vs.~low: 
$\beta=-0.62 [-6.18, 4.92]$). 
% JustDoIt results (direct action)
\HypETwoDoIt~was likewise supported within each cost condition: in the high cost condition (high~vs.~low: $\beta=-7.28 [-11.27,-3.29]$, low~vs.~no: $\beta=-24.55[-29.53,-19.60]$) and 
in the low cost condition (high~vs.~low: $\beta=-8.58 [-12.60, -4.48]$, low~vs.~no: $\beta=-46.34 [-55.98,-36.62]$).
% direct actions were more likely under low uncertainty than high uncertainty ($\beta=-7.28 [-11.27,-3.29]$), and likewise more likely under no uncertainty than under low uncertainty ($\beta=-24.55[-29.53,-19.60]$). %tns: can we make this transitive A > B > C more streamlined?
    %\begin{draftytext}
    %    JustDoIt was likewise supported within each cost condition.
    %    In the high cost condition, direct actions were more likely under no uncertainty than low [mathmath], which was in turn more likely than under high uncertainty [mathmath]
    % In the low cost condition...
    %\end{draftytext}
%Within the low cost condition (left facet of Figure \ref{ex:e2}), we found the same pattern for CQs (low~vs.~no: $\beta=18.60 [13.25, 24.00]$, high~vs.~low: 
%$\beta=-0.62 [-6.18, 4.92]$), 
%and
%the same predicted effect on direct action likelihood (low~vs.~no: $\beta=-46.34 [-55.98,-36.62]$; high~vs.~low: $\beta=-8.58 [-12.60, -4.48]$). 

Turning to cost, we also found smaller but credible effects. 
% Across uncertainty conditions, we observe that CQs are rated as credibly more likely in the high cost than in the low cost condition ($\beta=4.47 [0.38, 8.48]$). 
We confirmed \HypETwoHCQ:~CQs were more likely
when costs of acting incorrectly are higher ($\beta=4.47 [0.38, 8.48]$),
% We also find that direct actions are less likely in the high cost condition than in the low cost condition ($\beta=-6.60 [-9.69,-3.47]$), suggesting that participants are mildly sensitive to the cost of acting incorrectly.
and likewise, that direct actions become less likely, credibly supporting \HypETwoMessUp~($\beta=-6.60 [-9.69,-3.47]$). 
The remaining control reaction options of non-action, non-clarification question, and verbal acceptance were rated much less likely than CQ or direct action ($\beta=42.69[35.78, 49.35]$). 
While these options were occasionally used, their limited use reflects the predicted infelicity of responding to a directive with asking a question that fails to bear on the decision problem (non-clarification question) or doing nothing (non-action), with or without acknowledgment of the task (verbal acceptance). 
We also found that the effect of uncertainty (low~vs.~no uncertainty) was credibly larger on CQs than on non-clarification questions ($\beta=22.71 [15.84, 29.50]$), indicating that the manipulation of uncertainty does not appear to affect question-asking in general; it is only consequential for questions which bear on the DP at hand, namely, CQs. 
    %tns: love that we had space to include this sentence
    %km: <3
% \pt{hopefully there will be space for a sentence on reference vs manner vs deadline!}

Lastly, we found that referential (\textit{which}) CQs tend to have, on average, an appreciably lower likelihood rating than manner (\textit{how}) or deadline (\textit{when}) CQs. 
    %\ts{question commented here} 
    Exploratory by-type regression analyses suggest that the magnitude of the effects for CQs was driven by referential CQs, and qualitatively the same but marginally not credible for manner and deadline CQs; and while the gradient effects for direct actions were driven by deadline and manner contexts, a binarized effect was found in the referential contexts.
    This may be due to the typical domain sizes
%of each \textit{wh}-word: 
associated with each context type:
while domains of entities are often finite and constrained, domains of manner or time can be much more indeterminate, such that the expected reduction of uncertainty from a CQ may be lower.

\section{Computational Model}
\label{sec:model}
%\pt{clearer motivation of assumptions?}
As predicted by rational choice theory, and confirmed by our experimental results, participants in our experiments seemed to be sensitive to both uncertainty in the decision space and to the costs of possible actions.
The interaction of these factors can be captured using \textit{expected regret}, which is equivalent to the \textit{expected value of perfect information} \citep{RaiffaSchlaifer_1961:DecisionTheory}, and which reflects the expected difference between the expected utility \textit{before} seeking information and the expected utility \textit{after} eliminating all uncertainty.
We formulate these ideas in  a probabilistic model aimed to capture the empirical data from Experiment 1.

The decision problem is represented as a tuple $\langle G, P, R, U \rangle$, where $G = \{g_1, g_2\}$ are the two contextual goals of the questioner, probability distribution $P \in \Delta(G)$ captures the decision-maker's uncertainty about $G$, the set of available actions $R = \{r_{\textit{exh}}, r_{\textit{ms1}}, r_{\textit{ms2}}\}$ comprises all direct answers  (see Figure~\ref{fig:e1}, right), and the utility function $U: G \times R \xrightarrow{} \mathbb{R}$ describes the agent's payoff.
We assume that the four conditions of Experiment~1 differ in the degree of contextual uncertainty $\epsilon \in \{\epsilon_H, \epsilon_L\}$ for the high- and low-uncertainty conditions and in the cost associated with the exhaustive answer $\delta \in \{\delta_L, \delta_S\}$ for the large and small option spaces.
The resulting parameterized decision problem thereby becomes:
\begin{center}
    	\begin{tabular}{l|l|l|l|l}
		Feature & $P(g_i)$ & $U(r_{\textit{ms1}})$ & $U(r_{\textit{ms2}})$ & $U(r_{\textit{exh}})$ \\ \hline
		$g_1$ & $1-\epsilon$ & 1 & 0 & 1 - $\delta$  \\
		$g_2$ & $\epsilon$ &  0 & 1 &   1 - $\delta$  
	\end{tabular}
\end{center}

The core idea of the model is that, intuitively, an agent's reasoning is layered:
    faced with an uncertain decision context,
    an agent must first decide whether they have sufficient information to act.
        If none of the available actions are \emph{good enough} (i.e., if expected regret is high; cf.~\citealp{loomes1982regret, bourgeois2010regret}), the agent may opt to update the DP by seeking clarification, rather than simply taking one of the available options.
Formally, the agent \textit{reacts to} their contextual uncertainty, picking the clarification question $r_{\textit{cq}}$ with probability $P(r_{\textit{cq}})$, or otherwise \textit{acts under} uncertainty following a behavioral policy $\pi$:
    \begin{align*}
        P(r) = 
        \begin{cases}
            P(r_{\textit{cq}}) & \text{if } r = r_{\textit{cq}} \\
            (1 - P(r_{\textit{cq}})) \ \pi(r) & \text{if } r \in R\\
        \end{cases}
    \end{align*}%
Here, $\pi \in \Delta(R)$ is a soft-max policy with parameter $\alpha>0$:
    \begin{align*}
        \pi(r) &= \text{SoftMax}(\alpha \ \text{EU}(r)), \text{with} \\
        \text{EU}(r) &= \sum_{g \in G} P(g) \ \text{U}(g, r)
    \end{align*}
The probability $P(r_{\textit{cq}})$ is defined as a monotonic function of the expected regret of the \textit{best} action $r^* = \arg \max_{r} \pi(r)$ under the behavioral policy $\pi$. 
We use the logistic function $\text{Logistic}(x) = (1+\exp(-x))^{-1}$: 
\begin{align*}
    P(r_{\textit{cq}}) 
    &= \text{Logistic}(\tau \cdot (\text{ExpRegret}(r^*) - c )) 
    % \\ 
    % & = (1+\exp(-\tau \cdot \text{ExpRegret}(r^*) + c ))^{-1}
\end{align*}
where $\tau$ and $c$ are parameters for shape and location.  
Finally, $\text{ExpRegret}$ is defined as:
\begin{align*}
     \text{ExpRegret} (r^*) & = \sum_{g \in G} P(g) \cdot \text{Regret}(g, r^*) \\
\text{Regret}(g, r^*) & =  \arg \max_{r \in R} U(g, r) - U(g, r^*).
    \end{align*}

%\pt{please help me with feedback on whether this is how people report computational modeling results :'D. I feel like a primary school student writing something for the first time here.}

\paragraph{Bayesian Model Fitting} 
For a given tuple of model parameters $\theta = \langle \epsilon, \delta, \alpha, \tau, c \rangle$, the model's prediction of answer choices is $p^\theta = \langle P(r_{\textit{cq}}), P(r_{\textit{exh}}), P(r_{\textit{ms1}}), P(r_{\textit{ms2}}) \rangle$, which we take to be the (multinomial) predictor for the data from one of the four conditions from Experiment~1.
We use uninformative or flat priors:
$\delta_{L,S} \sim \text{Uniform}(0,1)$, 
$\epsilon_{L,H} \sim \text{Uniform}(0,0.5)$, 
$\tau \sim \text{Uniform}(0, 5)$, 
$c \sim \text{Uniform}(0, 1)$,
and $\alpha \sim \mathcal{N}(5,1)$ (truncated to be positive).
Without loss of generality, we assume that $g_1$ is the preferred goal in the low uncertainty condition, clamping $\epsilon \in [0, 0.5]$. 
We used Stan \citep{gelman2015stan} to obtain posterior parameter samples from four MCMC chains, each with 3000 warm-up and 4000 main samples.
Robustness of the inference was assessed by inspecting the usual indicators ($\hat{R}$ was $<1.01$, ESS above 2000).%

The estimates for cost and uncertainty parameters are as expected for the four experimental conditions 
    (large option space $\delta_L=0.32 [0.22, 0.43]$ was higher than $\delta_S=0.11 [0.00, 0.21]$ (small option space); 
    high uncertainty $\epsilon_H=0.49 [0.46, 0.50]$ was higher than $\epsilon_L=0.17 [0.05, 0.30]$ (low uncertainty)), 
confirming that, without a priori constraints, the data informs the model 
%in the predicted way.
as predicted.
Estimates for the logistic link function parameters were $\tau=3.60 [1.95, 5.00],~c= 0.18	[0.06,0.30]$. 
The posterior means and 95\% credible intervals of the posterior predictive distributions for each condition are shown in 
Figure~\ref{fig:e1} (black dots, CrIs). 
A visual posterior predictive check \citep{kruschke2010bayesian} suggests that the empirical proportions are within the credible interval of the proportions of 
CQs predicted by the model.
This is corroborated by a non-significant Bayesian posterior predictive $p$-value ($\textit{Bpppv} \approx 0.48$) obtained when using the (binomial) likelihood of CQ-choices as test statistic. 
However, Figure~\ref{fig:e1} also shows that the model does not match the human data perfectly. 
While the model predicts that there should be no preference for either mention-some answer in the high-uncertainty condition, participants seemed to have had a consistent bias towards choosing the option that was mentioned last in the description of the vignettes. 
Indeed, using the (multinomial) likelihood for the whole dataset we obtain $\textit{Bpppv} \approx 0.003$, suggesting that the model does not fully capture the variance in the data.

\begin{figure}
    \centering
    \includegraphics[width=\linewidth]{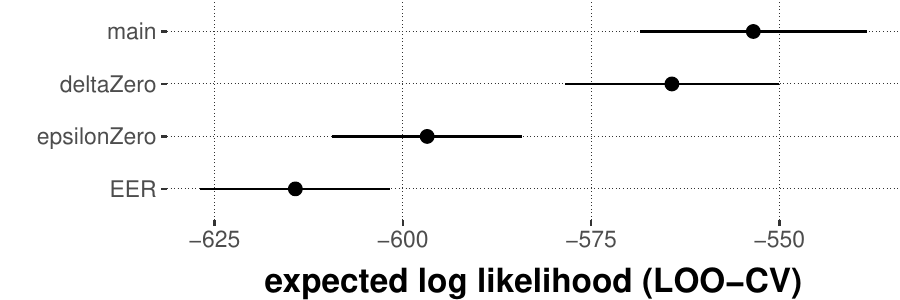}
    \caption{Estimates of expected log-likelihoods (dots) with standard errors (bars) for each 
    %of the four models (described in the main text). 
    model.
    Higher values are better.}
    \label{fig:loo-comparison}
\end{figure}

% \begin{figure}
%     \centering
%     \includegraphics[width=\linewidth]{latex/figs/posterior_pred_model_comp.png}
%     \caption{\pt{@MF: Instead / on top of the Bpppv's, as a last resort solution, we could use a plot like this (suggestions on better design are more than welcome!) to appeal at least to a visual posterior predictive comparison, because we see that the alternative models fail at least in some condition(s). What do you think?}\ts{reeeeally hard to track those dot-symbols, especially 4 vs. human}}
%     \label{fig:posterior-pred-comparison}
% \end{figure}

To further investigate whether the model's main functional ingredients are plausible against the background of the observed data, we compare it to three alternative models.
First, we consider an ablated model by setting $\delta=0$ to isolate the contribution of reasoning about cost. 
Second, to investigate the role of uncertainty about $G$, we ablate by setting $\epsilon = 0$.
Third, instead of using $\text{ExpRegret}(r^*)$, we use an expectation over the expected regret of all response options and pass it to a parameterized power law link function (EER model).  
Results from leave-one-out cross-validation are shown in Figure~\ref{fig:loo-comparison}.
A simple $z$-test suggests that the main model 
%considered here 
is indeed significantly better than the second best model with ablated $\delta$ ($p \approx 2.54 \cdot 10^{-8}$).
We conclude that our model is reasonably piloting a computational account of how humans may decide between 
reacting to 
%reducing
or acting under conversational uncertainty, suggesting that expected regret may be an adequate measure of 
%how human language confront \ts{???}
how people confront
the non-trivial trade-off between contextual uncertainty and action cost. 
%\pt{future work: our model is essentially a threshold model; alternative ways of deriving the threshold than expected regret (e.g., cost of action, uncertainty threshold) could be explored. We anticipate that these will not be able to capture the interaction where the threshold would be different in the conditions like the once we explored. Other pragmatic frameworks: RSA, game-theory?}

%\paragraph{Planned simulations / analyses}
%\begin{itemize}
%    \item prior predictive simulations, reporting whether predictions can be captured \pt{under uninformative priors? can we also manually set ``intuitive'' values for epsilon and delta and see what patterns are predicted? how exactly to quantify this?} 
%    \item \pt{parameter recovery? what exactly does that mean practically, and what does it give us beyond posterior predictive checks?}
%\end{itemize}
%\begin{itemize}
  %  \item total LLH of human data -40.39 [-43.62, -37.80] \pt{summing over the four conditions - is that correct?}
  %  \item alternative models' total LLH: (1) setting both epsilon to 0: -85.86 [-89.04, -83.49]; (2) setting both delta to zero: -53.19 [-56.29, -50.80]; exponential model from Michael (all Bpppv 0): -103.03 [-105.69, -101.13]
%\end{itemize}
%We assume that the utility of a mention-some answer listing, e.g., soft drinks when the true $g_{i^*}$ is soft drinks is 1, and 0 otherwise, and exhaustive answers have lower utility due to higher production cost, depending on the number of available options (Table~\ref{tab:payoffs}).

\section{Discussion}
Our work focused on answering the questions: when do agents act to reduce uncertainty---in particular, through communication, by asking CQs---and when do they act despite uncertainty, and what factors bear on that choice? 
Based on previous decision-theoretic work, we predicted that both the amount of 
decision-relevant %tns: are we distinguishing / did we ever define, the difference between uncertainty and decision-relevant uncertainty? isn't it clear that all of the uncertainty we're interested in is the kind which bears on the decision? or does this term just sound nice? :P
% pt: we said in the intro under point (i) that CQs are specifically sensitive to decision-relevant features, and yes, i do think it sounds nice^^ %tns: i had the same thought there, for what it's worth :P
uncertainty 
and the relative cost of available actions
factor into that decision, and that they interact.
That expectation was borne out in both experiments %we 
reported here: 
    Experiment 1 focused on linguistic responses to questions, and showed that agents are sensitive to both uncertainty and cost, asking more CQs under higher uncertainty---especially when providing an exhaustive answer was costly.
    Experiment 2 focused on the competition between linguistic and non-linguistic reactions to directives, and again demonstrated the same trade-off: %both uncertainty and cost affect how agents decide whether and when to ask CQs:
        participants were more likely to ask CQs when uncertain or %when 
        %mistakes of actions were more costly, while being less likely to take non-linguistic actions 
        when errors were more costly.
        %\pt{let's make this more concrete, talk about results for some of the H's} 
        %tns: potential paragraph break here? possible, not mandatory
        
        Experiment 2 revealed two distinct patterns with respect to uncertainty. 
        The rate of CQs showed a binarized pattern: credibly lower when there was no uncertainty, but no difference between non-zero uncertainty conditions, consistent with a threshold process where any uncertainty triggers consideration of clarification.
        The rate of direct actions, by contrast, showed a gradient pattern, declining with increased uncertainty across all three conditions, suggesting sensitivity to the degree of uncertainty when choosing among actions.
      %  This seeming inconsistency can be understood if we disentangle the question, `Should I ask a CQ?' from the question, `(Given that I'm taking a non-linguistic action,) What action should I take?'.
        This pattern is exactly in line with the layered structure of our computational model: %which first captures the agent's soft-thresholded decision if her current information is sufficient, ande then suggests a softmax-rational action selection. 
       % And indeed this is precisely the layered approach we encoded in our computational model: %\pt{@TS do you (strongly) think this should go here, as opposed to after the outlook to fitting the model to E2, as a speculation that the model should be able to fit that pattern? } 
        %tns: My thought was: [it's a key result from E2 > let's explain it > hey that segues nicely into talking about the layered model]
            %tns: my tendency would be: key results > interesting tidbits (less key results) > limitations > next steps/future directions   :)
            considering both the uncertainty around and relative costs of the actions available---modeled via \textit{expected regret}---, the agent first decides whether they have sufficient information for addressing the DP (in other words, whether to ask a CQ), and only afterwards 
            would %ask 
            decide
            which (non-CQ) action to take.
            The former decision procedure uses a (soft) threshold, 
                %tns: is it important that it's soft, here?
            %uses a binary measure of uncertainty, 
                %tns: I think I miss the parallelism here, but if you prefer it, so be it!
                % pt: i don't feel comfortable saying that it is straight up binary, because the agent picks a CQ with certain P (sampling), not deterministically once P > some t. Soft seems a way to express that, but I am actually rather in uncertain territory here. This is a quote from messages from Robert or Michael or both.
                % but it could make use of a binary measure/conception of uncertainty, even if the method by which it gets there is more complicated!   ¯\_(ツ)_/¯
            while the latter, ranking options by expected utility, uses a gradient measure.
            
Quantitative evaluation through Bayesian fitting of the model and comparisons to ablated model versions (without either uncertainty or cost) suggest that the expected regret-based model better explained the key interaction between uncertainty and cost in the human data. 
In other words, alternative models deriving the soft threshold for $P(r_{cq})$ from simpler uncertainty or cost based heuristics provided worse fit to human data.
However, the proposed model might not fully capture conditions where the rate of exhaustive answers was higher than the CQ rate. 
It also did not account for biased rates of mention-some answers, suggesting that factors beyond expected regret such as response noise or recency effects may also influence answer selection.

The experiments and model presented here make certain assumptions which are worth highlighting.
    Experiment 1 made use of a special kind of conditional (of the form `If $\langle$goal$\rangle$, $\langle$mention-some$\rangle$') to make clear the relationship between the response and the potential discourse goal \citep[cf.][]{Noh:1998-MetareprensationConditionals,Pittner:2000-Sprechaktbedingungen}.
        In theory, many more potential responses would be available to an agent (and relevant to the question) than were possible options in this experimental setup.
        A parallel experimental design using a free-choice production task might shed more light on which specific formulations speakers prefer.
    Similarly, Experiment 2 provided only one option for a non-clarificatory question (selected so as to remain on-topic without addressing the immediate DP); it is possible that speakers might have preferred other forms of non-clarificatory questions.
        That said, any such question would likely be interpreted in such a context as equally `non-compliant', a de facto challenge to the directive, carrying with it social costs and thus reduced utility.
    Finally, our model's calculated expected regret depends on the scale of the utilities of the possible actions, for which we assume particular values (informed by %prior work; 
    \citealp{hawkins2025relevant}); alternative assumptions about those utilities or their scale would lead to a different range of expected regret, which might require different priors for the parameterization of the logistic link function.

As developed thus far, our model handles linguistic responses of the type available in Experiment 1, for which relative production costs are reasonably estimable and comparable (listing eight drinks is more effortful than listing four).
We anticipate that the model should generalize to the setting of Experiment 2 as well, but this involves selecting among both linguistic and non-linguistic actions, which may have fundamentally different utility structures.
For instance, the cost of giving an overly long answer differs in kind from the cost of breaking down the wrong door: one wastes time, while the other may be irreversible. There is no obvious common currency for comparing these costs, and the same action (e.g., asking a CQ) may carry different relative value depending on the alternatives it competes against. 
Adapting the model to mixed action spaces would require principled ways of specifying utilities across modalities; we leave this to future work.

Several further avenues remain for future work.
We have focused on the costs of \emph{not} asking (namely risking error from acting under certainty).
But there are also costs \emph{to} asking: CQs take time and effort, delay action, and may carry social costs such as appearing incompetent to a superior. 
While the parameter $c$ in our model could be interpreted as the cost of asking, we did not explicitly manipulate this; future experiments might do so through time pressure or social context. 

More broadly, our expected regret-based framework offers a unified approach to studying how agents navigate the tension between acting and inquiring. 
Clarification questions represent a distinctively human strategy for managing uncertainty, one that balances epistemic caution against the pressure to act. Understanding when and why people ask is a step toward understanding how humans coordinate action through language, and toward building systems that can do the same.

\section{Acknowledgments}
KM is funded by the Deutsche Forschungsgemeinschaft (DFG, German Research Foundation) under project ID 547376651.
TS and MF are funded by the Deutsche Forschungsgemeinschaft (DFG, German Research Foundation) under project ID 538184825 (CRC 1718).
MF is a member of the Machine Learning Cluster of Excellence at University of T\"ubingen, EXC number 2064/2 – Project number 39072764.

\printbibliography

\end{document}